\documentclass[lettersize,journal]{IEEEtran}
\usepackage{amsmath,amsfonts}
\usepackage{algorithmic}
\usepackage{algorithm}
\usepackage{array}
\usepackage[caption=false,font=normalsize,labelfont=sf,textfont=sf]{subfig}
\usepackage{textcomp}
\usepackage{stfloats}
\usepackage{url}
\usepackage{verbatim}
\usepackage{graphicx}
\usepackage{cite}
\begin{document}

\title{
A Task-Agnostic Control Strategy for Dynamic Assistance with Pneumatically Actuated Soft Exosuits}

\author{Anoush Sepehri, Zachary Huang, Raymond de Callafon, Michael T. Tolley, Tania K. Morimoto
\thanks{All authors are with the Department of Mechanical and Aerospace Engineering, University of California San Diego, San Diego, CA 92093, USA (Corresponding email: tamorimoto@ucsd.edu)}
}



\maketitle

\begin{abstract}

Pneumatic artificial muscles have provided new opportunities to develop upper-extremity soft exosuits for rehabilitation, augmentation, and assisted daily living. However, the complex dynamics and limited bandwidth of these actuators has made providing responsive assistance based on user intention a longstanding challenge. In this work, we present an inverse-plant control strategy for pneumatically actuated soft exosuits that only relies on kinematic sensing for task-agnostic and dynamic assistance during daily living. We model the human-robot system using a Hammerstein dynamic model, consisting of a Preisach hysteresis model and a linear time-invariant filter, to capture the static and dynamic behavior of the system. We personalize our model to each user using 140~s of data and approximate an inverse to integrate into our control loop. When evaluated on a test rig that emulated a soft assistive exosuit for the wrist,  our controller reduced the interaction torque by up to 73\% and the activation of key flexor and extensor muscles by up to 47\% relative to the condition with no assistance for speeds ranging from 8\textdegree/s to 120\textdegree/s. Overall, this work presents a control strategy that can provide task-agnostic, dynamic assistance with pneumatically actuated soft exosuits without the need for physiological or force sensors to interpret user intention.

\end{abstract}


\section{Introduction}

Neurological injuries, such as a stroke or spinal cord injury, are some of the leading causes of upper limb motor impairment in the United States \cite{ma2014incidence}. Motor impairments, such as weakness and paralysis due to these injuries, can lead to limited use of the injured limb, resulting in elevated joint stiffness, spasticity, and ultimately adapted non-use that hinders quality of life~\cite{andrews1979sroke}. Access to frequent and intense rehabilitation has been shown to improve recovery, however, current programs are costly and require one-on-one therapy sessions, limiting accessibility to the general public \cite{devittori2024unsupervised}.

To increase access to rehabilitation therapy after a neurological injury, researchers have developed robotic rehabilitation systems for both the upper and lower-extremities \cite{devittori2024unsupervised}. Starting with grounded robotic manipulators such as the MIT-MANUS \cite{krebs1998robot}, researchers have developed more complex robotic systems that directly interface with the user's joints and limbs, eventually leading to wearable and untethered exoskeletons that are powered by hydraulic actuators \cite{zoss2006biomechanical} and electromechanical motors \cite{perry2007upper}. As these exoskeletons became more advanced, researchers began exploring applications beyond rehabilitation, such as for human augmentation to reduce metabolic cost of walking \cite{siviy2023opportunities} or to assist with activities of daily living \cite{ochieze2023wearable}. While rigid exoskeletons can apply precise torques and forces to the body, their materials, actuators, and backend electronics result in bulky systems that are cumbersome and uncomfortable to wear \cite{ochieze2023wearable}. Additionally, misalignment between the exoskeleton and the user's joints can lead to undesirable interaction forces and discomfort \cite{ochieze2023wearable} that must be accounted for with complex self-alignment mechanisms~\cite{chen2024systematic}. Combined with the high cost of these devices \cite{gorgey2018robotic}, rigid exoskeletons have had limited adoption outside of rehabilitation facilities and specialized applications.

To address these challenges, researchers have begun to design wearable robots using soft and compliant materials \cite{siviy2023opportunities, ochieze2023wearable}. These robots, often referred to as exosuits, can accommodate natural body motions, compensate for joint misalignment, and resemble clothing that is already familiar to the user. Some of the most prevalent soft assistive exosuits are cable-driven and actuated using motors for gait assistance \cite{yang2022soft}, hand rehabilitation \cite{kim2025exo}, and lumbar support \cite{li2021design}. More recently, researchers have explored soft actuation methods, most notably pneumatic artificial muscles \cite{sepehri2024soft, yumbla2025personalized, proietti2023restoring, park2014design, schaffer2024soft, polygerinos2015emg, nassour2021soft}. These actuators are lightweight, possess a high force-to-weight ratio, and have inherent compliance for safe human-robot interaction that makes them especially attractive for wearable applications.

While there has been rapid progress in the design of soft exosuits, controlling these devices is an ongoing challenge. Most existing exosuits demonstrate their functionality through open-loop control or closed-loop tracking of predefined reference trajectories \cite{park2014design, schaffer2024soft}. While appropriate for demonstrating the functionality of the device, they do not account for a user's intentions during real-world deployment. For soft assistive exosuits to be successfully integrated into daily living, they must be capable of reliably estimating user intent and delivering responsive, effective assistance for a wide range of scenarios relevant during daily living activities.

\subsection{Existing Intent-Based Control Strategies}

Intent-based control strategies can be separated into two classes. The first class uses physiological sensing, most commonly through surface electromyography (sEMG), to detect human intention. Researchers have used sEMG for proportional myoelectric control \cite{lenzi2012intention, peng2024improving} or to classify actions such as flexing and extending the fingers \cite{polygerinos2015emg}. Recently, researchers have used EMG-driven musculoskeletal modeling to estimate the force generated by the targeted muscle to determine the level of assistance to generate from the robot \cite{lotti2020adaptive}. While EMG-driven musculoskeletal modeling has shown impressive results, precise torque estimation requires time consuming user calibration sessions and precise placement of the sEMG sensors on the muscle belly to get strong one-to-one signals of muscle activity. Furthermore, the sensors used in these works cost thousands of dollars, and changes in electrode contact, impedance drift, and poor robustness to sweat make it challenging to use such sensing modalities over long time horizons \cite{ferdousi2025complicacy, schonhaut2025emg}.

In parallel with physiological sensing, researcher have investigated a second class of intent-based control strategies that rely on interaction forces and kinematic data from the exosuit. For example, researchers have developed a control scheme that used a dynamic model of the limb to estimate the interaction torque from the user as a proxy for intention to control cable-driven exosuits for the elbow and wrist \cite{xiloyannis2019physiological, chiaradia2021assistive}.  While their control method had similar performance to model-based myoelectric control \cite{lotti2022myoelectric}, there were still a number of limitations. The exosuits required a load cell in line with the actuator to measure the torque applied to the joint, therefore increasing the system's complexity. Furthermore, the dynamic models used in these works only considered linear dynamic models for inertia and viscous damping of the joint \cite{xiloyannis2019physiological} or only the effects of gravity when holding a mass \cite{chiaradia2021assistive}. These assumptions may not be optimal when stiffness dominates the impedance, such as for the wrist \cite{charles2011dynamics}, and when nonlinear properties are not negligible. Furthermore, the authors tuned the parameters of their models empirically or based on anthropometric estimates instead of identifying optimal parameters based on experimental data from the user.

Pneumatic artificial muscles possess intrinsic compliance, meaning any interaction forces generated by the user on the robot leads to changes in the joint position. As a result, researchers have explored methods to detect user intent strictly based on kinematic data from the exosuit. This is especially valuable because it would eliminate the need for extra sensors (i.e., force or physiological sensing) during deployment. Researchers have developed controllers where the reference pressure to the actuators changed based on changes in the position \cite{ferroni2025soft} or measured pressure of the actuators \cite{realmuto2022assisting}. Others have developed a kinematic state machine to control when to inflate and deflate the actuators \cite{zhou2021kinematics, zhou2024portable}. These control strategies, however, provided discrete levels of assistance which may not be optimal for highly dynamic activities, as suggested by feedback from users \cite{zhou2024portable}. To provide continuous levels of assistance, researchers have developed gravity compensation techniques by relating the measured joint angle to the desired pressure for the actuator, however, they ignored all dynamic properties in their system \cite{proietti2021sensing, proietti2023restoring, mccann2025body}. 

\subsection{Soft Actuators versus Electromechanical Actuators}
While soft actuators have a number of properties that make them promising for assistive exosuits, they suffer in terms of controllability, efficiency, and performance when compared to motors \cite{chen2026call, heisser2026codevelopment}. Pneumatic artificial muscles have complex nonlinear behavior and are generally only suitable for quasi-static and low-speed applications due to their limited bandwidth. For example, a shoulder exosuit with unfolding textile pneumatic actuators required up to 1.8~s to deploy robotic assistance due to air flow dynamics \cite{proietti2023restoring}. McCann \textit{et. al.} developed a hysteresis model to capture the nonlinear and hysteretic behavior in a soft assistive exosuit for the shoulder, however, they only evaluated their model at quasi-static speeds $(<3^\circ\text{/s})$ to avoid any dynamic effects from influencing their model \cite{mccann2025body}. In general, the low bandwidth of these actuators introduce delays between the control signal and the assistance applied to the body, limiting performance \cite{young2017biomechanical, yumbla2025personalized}. To improve the responsiveness of such exosuits, Arnold \textit{et al.} developed a machine learning-based intent detection model to offset the input angles to an inverse hysteresis model \cite{mccann2025body} to anticipate future movement \cite{arnold2025personalized}. The intent detection model, however, required both kinematic and force sensing along the upper arm, had a lengthy calibration period $(\sim17~\text{min})$, and still had slow actuation speeds $(500~\text{ms})$ due to inference delays of the model. Realmuto \textit{et al.} \cite{realmuto2022assisting} used a bang-bang pressure controller and a pneumatic reservoir maintained at 865 kPa (125 psi), approximately twice the maximum operating pressure of the actuators, to maximize the system's flow capacity. Such a large reservoir, however, limits portability and can introduce safety concerns for human-robot applications. Furthermore, the combination of a bang-bang controller and high supply pressure led to non-smooth assistance, resulting in increased tracking errors during a marker following task.

\subsection{Contributions}

To achieve responsive and continuous assistance with pneumatically actuated soft exosuits, we present a inverse-plant control strategy that can provide assistance for the speeds and frequencies relevant to daily living activities without requiring any force or physiological sensing. The contributions of our work are as follows:

\begin{enumerate}
    \item \textbf{Theoretical Analysis and Simulation Validation:} We analyze the stability and robustness of a control strategy based on the inverse-plant method. We also present a simulation of our controller for a single degree-of-freedom joint to demonstrate its benefits compared to existing methods.
    
    \item \textbf{Inverse-Plant Model and Controller Design:} We present an inverse-plant controller based on the Hammerstein dynamic model and introduce an efficient data collection procedure to capture the static and dynamic behavior of the human-robot system. We then derive an inverse approximation of our model to integrate into a feedback loop for personalized assistance based solely on kinematic data from the robot.

    \item \textbf{Experimental Validation:} We experimentally validated our controller over different speeds, frequencies, and scenarios relevant for daily living using a test rig that emulates a soft assistive exosuit for flexion and extension assistance of the wrist.
\end{enumerate}

The remainder of this paper is organized as follows. In Section II we introduce the principles of our control strategy and provide a simulation highlight the benefits of our controller. In Section III we describe our test rig for emulating a soft assistive exosuit for the wrist. We also present our method for modeling the forward and inverse dynamics of the human–robot system using the Hammerstein dynamic model. In Section IV we evaluate the accuracy of our modeling approach against common models seen in literature. Lastly, in Section V we evaluate our inverse-plant control method on users with no motor impairments during several virtual marker following tasks.

\section{Controller Principle}

\subsection{Inverse-plant Controller Design}

\begin{figure}[t]
\centerline{\includegraphics[width=0.49\textwidth]{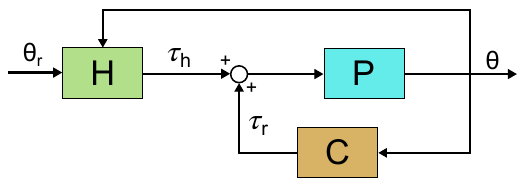}}
\vspace{-5pt}
\captionsetup{labelformat=empty}
\caption{An idealized block diagram of the human-robot system (P), the proposed, positive feedback controller (C), and the human’s internal motor controller (H). The controller has access to the kinematic state of the robot $(\theta)$ and generates assistive torques $(\tau_r$) to augment the user's effort $(\tau_h$)}
\label{fig1}
\vspace{-15pt}
\end{figure}

A human wearing a soft assistive exosuit (i.e., the human-robot system) can be modeled simply as a plant $(P)$ that represents the impedance of the joint combined with the device, an internal controller of the human $(H)$ that maps desired joint positions $(\theta_r)$ to muscle activation and biological torque $(\tau_h)$, and the controller on the exosuit $(C)$ which commands assistive torques $(\tau_r)$ that are applied to the joint (see Fig.~\ref{fig1}). We limit our controller to only having access to kinematic information of the robot for detecting human intention and calculating an assistive torque $(\tau_r)$ to supplement the user's effort $(\tau_h)$. The transfer function that represents the amount of effort from the user to follow a trajectory ($\theta$) is defined as follows:
\begin{equation}
\frac{\tau_h}{\theta} = T(s) = \frac{1-PC}{P}
\label{eq1}
\end{equation}

Our goal is to minimize the magnitude of Eq.~\ref{eq1} over the frequency band relevant for daily living using our controller $(C)$. Let us define our controller as the inverse of our plant dynamics $(P)$ scaled by $\alpha$ \cite{kazerooni2005control}:
\begin{equation}
C = (1-\alpha)P^{-1} \text{ for } \alpha = (0, 1]
\label{eq2}
\end{equation}

\noindent where smaller values of $\alpha$ correspond to higher levels of assistance from the controller. Then Eq.~\ref{eq1} simplifies to the following:
\begin{equation}
    T(s) = \frac{\alpha}{P}    
\end{equation}

By designing our controller in this way, we can minimize the effort necessary from the user by a scaling factor $\alpha$ for a trajectory defined by $\theta$. This approach has two benefits compared to the state-of-the-art control strategies for upper extremity soft assistive exosuits: (1) it provides continuous levels of assistance for any arbitrary trajectory, and (2) it does not require any extra sensors in the control loop to measure muscular activity or interaction forces (Note: in testing this controller, we use force and EMG sensors for experimental evaluation, but not as inputs to the controller, see Sec.~\ref{sec:emulator}).

\subsection{Stability and Robustness to Plant Uncertainty}

We now evaluate the properties of our proposed control strategy when subject to plant uncertainties. The absolute sensitivity of Eq.~\ref{eq1} to modeling uncertainty is as follows:
\begin{equation}
    \frac{\Delta T}{T} = \frac{\partial T}{\partial P}\frac{P}{T}\frac{\Delta P}{P} = \left(\frac{1}{1-PC}\right)\frac{\Delta P}{P}
    \label{eq3}
\end{equation}

When $C$ is defined as in Eq.~\ref{eq2}, the sensitivity function simplifies to the following:
\begin{equation}
    \frac{\Delta T}{T} = \left(\frac{1}{\alpha}\right)\frac{\Delta P}{P}
    \label{eq4}
\end{equation}

For very aggressive controllers $(\alpha \to 0)$, $T(s)$ becomes very sensitive to plant uncertainty. Consequently, more accurate models of the human-robot plant lead to more effective and reliable assistance.

Existing literature has suggested that humans coordinate movement through feedforward, $F(\theta_r)$, and feedback, $B(\theta_r-\theta)$ control ($H(s) = F(\theta_r) + B(\theta_r - \theta)$) where $\theta_r$ is the desired position which is decided by the user based on the task. The feedforward component is based on an internal dynamic model of their limb  and the feedback component can be modeled as a proportional-integral-derivative (PID) controller \cite{wolpert1998multiple, kawato1999internal}. The transfer function mapping the desired position $\theta_r$ to the actual position $\theta$ is as follows:
\begin{equation}
    \frac{\theta}{\theta_r} = G(s) = \frac{FP + BP}{1-PC+BP}
    \label{eq5}
\end{equation}

The stability of our system is defined by the characteristic equation ${1-PC+BP}$. To simplify our analysis, we assumed that the human's internal feedback controller was a proportional controller with gain $K_p$. We also assumed that the human-robot system $(P)$ can be modeled as a dynamic system with some stiffness $(k)$ and damping $(b)$ \cite{charles2011dynamics}. Due to modeling errors when identifying our controller, we include an uncertainty parameter $(\lambda)$ as part of our plant.
\begin{equation}
    P = \frac{1}{\lambda(bs+k)}   
    \label{eq6}
\end{equation}

\noindent Our characteristic equation then becomes the following:
\begin{equation}
    1-PC+BP = 1 - \frac{(1-\alpha)}{\lambda} + \frac{K_p}{\lambda(bs+k)}
    \label{eq7}
\end{equation}

\noindent where the zeros are as follows:
\begin{equation}
    s = \frac{-((\lambda + \alpha -1)^{-1}K_p+k)}{b}
    \label{eq8}    
\end{equation}

For no plant uncertainty $(\lambda = 1)$,  the zeros are in the LHP for $\alpha, K_p, k, b > 0 $. With plant uncertainty, however, stability is guaranteed only when  $\lambda + \alpha \geq 1$ for $K_p,~k,~b > 0$. This observation provides several important insights: (1) The performance of our controller (and consequently the level of assistance) is directly influenced by the fidelity of our identified plant. The more accurate our identified plant is $(\lambda \to 1)$, the more aggressive we can design our controller $(\alpha \to 0)$. (2) Instability only occurs when the identified parameters for the controller overestimate the true parameters in the plant $(\lambda < 1)$. Humans modulate the impedance of their joints depending on the scenario or task \cite{selen2006impedance}, thus implying the most accurate dynamic model would be time-varying. However, if we identify a model that captures the lower bound of the plant parameters (i.e., when the user is fully relaxed), any instabilities that could arise would naturally be stabilized by the user contracting or co-contracting their muscles to increase joint impedance \cite{winters1988analysis}. We believe the stabilizing capabilities of the user in the control loop, along with the inherent compliance and safety that comes with a soft robot, considerably reduces the practical drawbacks that limited robustness has when implementing our control strategy.

\subsection{Simulation Evaluation}

\begin{figure}[t]
\centerline{\includegraphics[width=0.49\textwidth]{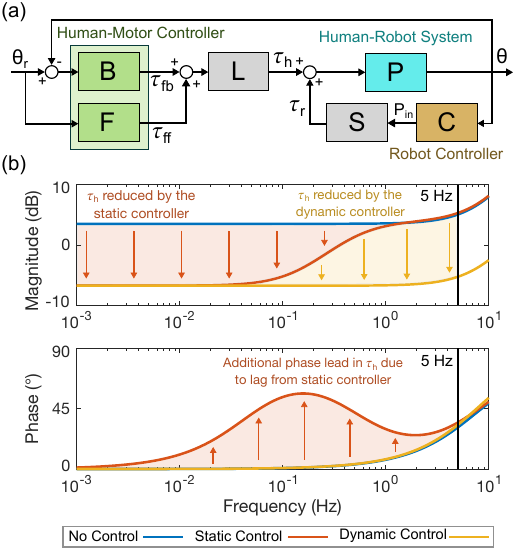}}
\vspace{-5pt}
\captionsetup{labelformat=empty}
\caption{Simulation results to evaluate our proposed control strategy. (a) We expanded on the block diagram in Fig.~\ref{fig1} to include muscle dynamics and coupling dynamics between the human and the robot. (b) While existing static control techniques can provide effective assistance (i.e., a reduction in the effort, $\tau_h$) in quasi-static scenarios, they fail to provide meaningful assistance at higher frequencies necessary for dynamic movements.}
\label{fig2}
\vspace{-15pt}
\end{figure}

We evaluated our inverse-plant control strategy in simulation for an idealized, single degree-of-freedom joint. We extended our analysis from the preceding section by including two additional components for our simulation (see Fig.~\ref{fig2}a). First, we incorporated a low pass filter $(L)$ with a rise time of  60~ms to account for muscle dynamics and sensory delays in the neuromuscular system \cite{crevecoeur2013priors}. We also included a low pass filter $(S)$ with a cutoff frequency of 0.5 Hz to capture the air flow and soft coupling dynamics between the human and robot to map the reference pressure to an assistive torque $(\tau_r)$ \cite{mosadegh2014pneumatic}. We modeled the human-robot system $(P)$ as a linear spring-damper system $(k = 1.5,~b = 0.03)$ \cite{charles2011dynamics}. The user’s internal motor controller was a proportional gain feedback controller $(K_p = 50)$ combined with an inverse-plant feedforward controller \cite{wolpert1998multiple, kawato1999internal}. We assumed that the user had a good internal representation of their joint impedance to generate a perfect feedforward controller. 

The transfer function mapping the desired position $(\theta_r)$ to the human effort $(\tau_h)$ can be represented as follows:

\begin{equation}
    \frac{\tau_h}{\theta_r} = \frac{(FL + BL)(1-PCS)}{1-PCS+BLP}
    \label{eq10}
\end{equation}

The objective was to design a controller $(C)$ that minimized Eq.~\ref{eq10} over the frequency band of interest.

We simulated two different inverse-plant control strategies: a static controller that did not incorporate any dynamics and a dynamic controller that did incorporate dynamics, to evaluate their effects on Eq.~\ref{eq10} (see Fig.~\ref{fig2}b). We also simulated a no control condition $(C=0)$ to function as the baseline.

For the static controller, we implemented an inverse-plant model that compensated for the stiffness of the human joint. This case replicated the gravity compensation techniques seen in previous literature to support arm abduction \cite{proietti2021sensing}. We assumed that we could successfully identify the stiffness parameter in the plant $(P)$ and designed our controller as follows:
\begin{equation}
    C = (1-\alpha)k
    \label{eq11}
\end{equation}

\noindent where $k$ is the stiffness of the joint, and $\alpha = 0.3$ to scale the level of assistance. The static controller minimized human effort in the quasi-static frequencies, however, as the frequency increased, the controller became less effective due to the dynamics of the human-robot system not captured by the controller (see Fig.~\ref{fig2}b). At approximately 1 Hz, the controller had negligible benefits when compared to the baseline case. The controller also introduced phase lead into the user's motor patterns due to the lag from the assistance that had to be compensated for by the user. While this controller may be effective for supporting static positions and quasi-static movements, it would not be effective for dynamic movements which are typically seen in joints such as the wrist and fingers. 

For the dynamic control case, we identified a model for the plant (P) and dynamics of the robot (S). Our controller was then modeled as follows with $\alpha = 0.3$:
\begin{equation}
    C =(1-\alpha)(PS)^{-1} 
    \label{eq12}
\end{equation}

By identifying a model of the human-robot system that incorporated dynamics, we achieved effective assistance over a broader range of frequencies (see Fig.~\ref{fig2}b) with minimal influence on the user's motor patterns. Approximately 75\% of the spectral density in wrist motion during daily living activities is captured at frequencies up to 5 Hz \cite{mann1989frequency}. Consequently, a control strategy that can provide effective assistance at these frequencies would be especially beneficial to achieve seamless human-in-the-loop assistance during daily activities.

\section{Materials and Methods}

Based on our analysis, designing an effective inverse-plant controller for a soft assistive exosuit can be simplified to identifying an invertible mathematical model with enough fidelity to capture the principal dynamics of the human-robot system. This results in several practical challenges: (1) the dynamics of joint motion are complex and further complicated due to the nonlinear characteristics of the pneumatic artificial muscles and coupling between the human and the exosuit, (2) data-driven modeling techniques must be very efficient to ensure a brief calibration period per user, and (3) many linear, nonlinear, and black box models do not have an analytical inverse that is stable and causal. We focused the remainder of this work on addressing these challenges.

\begin{figure*}[t]
\centerline{\includegraphics[width=0.98\textwidth]{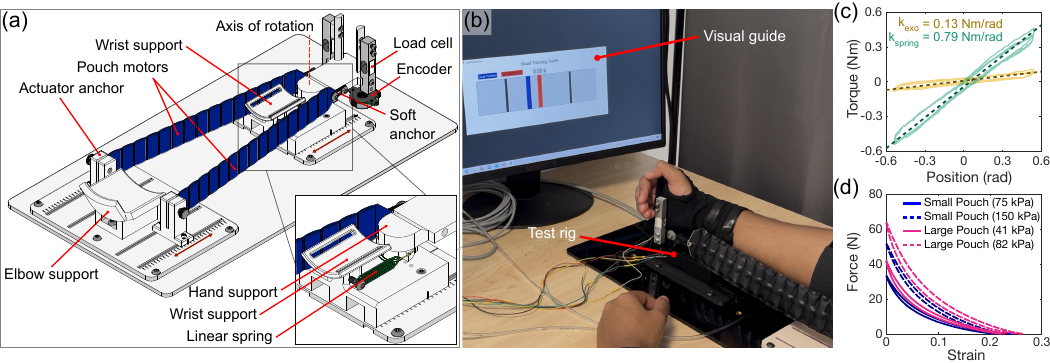}}
\vspace{-5pt}
\captionsetup{labelformat=empty}
\caption{Overview of the test rig for flexion and extension assistance for the wrist. (a) The test rig consisted of two pouch motors in parallel with the flexor and extensor muscles of the wrist and linear springs to simulate elevated joint stiffness after a neurological injury. The wrist support and anchors for the proximal end of the pouch motors can be adjusted to account for anatomical differences between users. (b) The experimental test setup consisted of the test rig and a visual guide for the marker following tasks. (c) The passive stiffness of the test rig with $(k_{spring})$ and without $(k_{exo})$ the linear springs to simulate elevated joint stiffness. The stiffness without the springs was due to the soft coupling to the pouch motors. (d) Force-strain characterization results for the small and large pouch motors used in this work.}
\label{fig3}
\vspace{-15pt}
\end{figure*}

\subsection{Soft Exosuit Emulator}
\label{sec:emulator}

Researchers have developed test rigs, frequently referred to as emulators, to evaluate different control strategies for multi-joint lower extremity exoskeletons \cite{bryan2021hip}, ankle prosthetics \cite{caputo2014universal}, and tremor suppression for the wrist~\cite{shomron2025robotic}. These emulators incorporated off-board power and actuation systems, and allowed researchers to rapidly explore different control strategies without the need for product-like prototypes that can require years of development. Similarly, we decided to develop a test rig that captured the most critical aspects of a soft assistive exosuit to comprehensively evaluate our control strategy. 

Our test rig consisted of three major components (see Fig.~\ref{fig3}(a)): (1) an antagonistic pair of pouch motors that were aligned with the flexor and extensor muscles of the wrist, (2) an elbow support with anchoring points for the proximal end of the pouch motors, and (3) a hand support with an integrated absolute position encoder, anchoring points for the distal end of the pouch motors, load cells, and linear springs to simulate elevated joint stiffness after a stroke.

We chose pneumatically actuated pouch motors for our emulator due to their high force output, large contraction ratios, and simple manufacturing procedure \cite{niiyama2015pouch}. We heat sealed TPU coated ripstop nylon (190 GSM Ripstop Fabric, TPU both sides, DIYpackraft) together with a heat resistant mylar layer to define the pouch geometries. We fabricated a small and large version of our pouch motors to evaluate on our modeling strategy when subject to exosuits with different air flow dynamics. The geometry of the pouches on the small actuator were 30.0~mm by 12.0~mm, while the geometry of the pouches on the large actuator were 25\% larger (37.5~mm by 15.0~mm).  We spaced each pouch 2.5~mm apart in the actuator to maximize the density of the pouches while also ensuring the  seams were wide enough to prevent delamination during inflation. We heat sealed the ripstop nylon using a heat press for 30 seconds at 370°F. To ensure equal heat transfer to both sides of the nylon, we flipped the pouch motors and conducted a final sealing for 30 seconds at 370°F again. The proximal end of the pouch motors were rigidly attached to anchors on the elbow support. The distal end of the pouch motors were attached to a textile loop to simulate the soft coupling when mounting the actuators to a wearable soft interface.

The elbow support was 3D printed using Polylactic Acid (X1C, Bambu Labs) and included velcro straps to hold the user's limb in place. The hand support was also 3D printed and included an absolute position encoder (AMT222B-V, Mouser Electronics) for kinematic feedback. We integrated two load cells into the hand support that secured the user's hand and measured the interaction forces between the user and the test rig during operation. We also integrated two linear springs (9044K253, McMaster-Carr) that wrapped around the axis of rotation of the wrist to simulate elevated joint stiffness when testing on individuals with no history of motor impairment. We selected the springs based on literature that studied the stiffness of patients wrists after a stroke \cite{wang2017neural}. 

The backend electronics consisted of two electro-pneumatic regulators with integrated pressure sensors (VEAB-L-26-D9-Q4-V2-1R1, Festo) that controlled the pressure to the pouch motors, a microcontroller (UNO, Arduino) that controlled the regulators using a digital-analog-converter (MCP4725, Adafruit) and sampled the pressure sensors, position encoder, and load cells, and a host computer that ran the control loop at 100 Hz and a graphical user interface at 30 Hz for a marker following task (see Fig.~\ref{fig3}(b)). Although our emulator consisted of two actuators with two independent inputs, they were operated mutually exclusively (i.e., both actuators were never pressurized simultaneously). We defined an abstract input signal where the sign defined which actuator to inflate and the magnitude represented the desired pressure. By defining our input signal in this manner, we simplified our system to a single-input and single-output (SISO) model \cite{park2014design}. 

\subsection{System Modeling and Identification}

Standard linear system identification techniques can provide reliable models from experimental data \cite{ljung}. However, assuming a soft assistive exosuit can be approximated as linear would significantly limit model accuracy and consequently the quality of assistance. Instead, characterizing the nonlinear components in our system based on existing knowledge and compensating for them allowed us to linearize the residual dynamics. By doing this, we directly exploited the robustness, interpretability, and sample efficiency of linear identification methods without neglecting the nonlinear properties of our system. We accomplished this by modeling the human-robot system as a nonlinear map followed by a linear dynamic filter (i.e., a Hammerstein dynamic model \cite{schoukens2017identification}, see Fig.~\ref{fig4}(a)i).

\subsubsection{Nonlinear Static Map}

Joint stiffness, hysteresis, and complex pressure-torque relationships contribute substantially to the overall nonlinearity of soft assistive exosuits that use textile pneumatic actuators \cite{mccann2025body}. As a result, we modeled these nonlinear properties using a Preisach hysteresis model. The Preisach hysteresis model captures hysteresis as a superposition of weighted, rate-independent delay operators (i.e., hysterons) \cite{mayergoyz1986mathematical}. The Everett contour can efficiently encode one-to-one mappings for clockwise and counter-clockwise hysteresis \cite{mayergoyz1986mathematical}, thus allowing us to efficiently estimate the forward and inverse hysteresis relationship. The Preisach hysteresis model has been shown to be effective for compensating for the hysteresis in supercoiled polymer artificial muscles \cite{zhang2017modeling}, soft-magnetic materials \cite{daniels2023everett}, and more recently, soft assistive exosuits for shoulder assistance \cite{mccann2025body}.
 
We collected concentric reversal curves of the human-robot system to identify the Everett contour (see the first block in Fig.~\ref{fig4}(a)i). We captured four concentric hysteresis loops (the major loop and three minor loops) based on the maximum operating pressures calibrated for each user. Each loop was collected over 20 seconds (0.05 Hz) to minimize any dynamic effects of the human-robot system from influencing the data. We then processed the data by histogram filtering the signal, ensuring endpoint alignment, and enforcing monotonicity in the loading and unloading curves \cite{mccann2025body}. After processing the data, we fit each loading and unloading curve using a smoothing spline with end-point constraints and uniformly sampled the result for an equal distribution of points on the Everett contour. We linearly interpolated between the points on the Everett contour to create a continuous surface for prediction.

\subsubsection{Linear Dynamic Calibration}
We identified a discrete-time transfer function $P(z)$ to capture the residual dynamics of the human-robot system (see $P(z)$ in Fig.~\ref{fig4}(a)(i)). We assumed the residual dynamics $P(z)$ could be parametrized by a second-order transfer function with one finite (discrete-time) zero to account for sampling dynamics according to
\begin{equation}
    P(z,\lambda) = \frac{b_0 +b_1z^{-1}}{1+a_1z^{-1}+a_2z^{-2}}
    \label{eq15}
\end{equation}
\noindent where $\lambda = [b_0 ~ b_1 ~ a_1 ~a_2]$ contained the parameters to be estimated. To identify $\lambda$, we used a sinusoidal chirp signal with frequencies up to 10 Hz that were logarithmically spaced over 60 seconds. We computed the value of $\lambda$ by minimizing the Least-Squares Error (LSE) of $e(t,\lambda)$ in the output-error model
\begin{equation}
    e(t,\lambda) = \theta(t) - \theta(t,\lambda),~~ \theta(t,\lambda) = P(z,\lambda)\tilde{u}_r(t)
    \label{eq13}
\end{equation}
\noindent where $e(t,\lambda)$ was the error between the measured output $\theta(t)$ and predicted output $\theta(t,\lambda)$ and $\tilde{u}_r(t)$ was the reference pressure input $u_r(t)$ after being remapped using the nonlinear static map from the preceding section. We used a Gauss-Newton iterative solver to perform the non-convex minimization of the LSE of $e(t,\lambda)$ using the Matlab System Identification Toolbox. The end result was a dynamic filter $P(z,\lambda)$ with complex-valued (discrete-time) poles and a single real-value zero, capturing the residual dynamics of the human-robot system useful for controller design.

\subsection{Controller Design}

\begin{figure}[t]
\centerline{\includegraphics[width=0.5\textwidth]{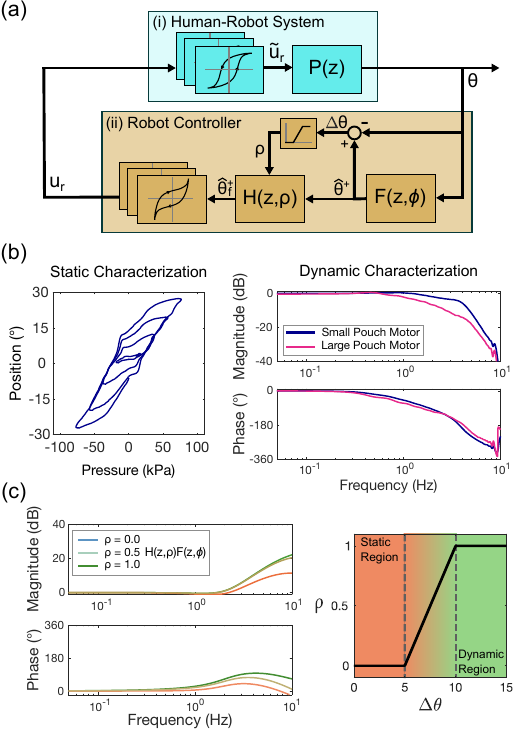}}
\vspace{-5pt}
\captionsetup{labelformat=empty}
\caption{The inverse-plant control strategy developed in this work (a) The (i) human-robot system represented as a Hammerstein dynamic model with the (ii) robot controller. (b) Experimental characterization results for a representative user to identify the parameters in the Hammerstein dynamic model. Left: The hysteresis properties of the human-robot system that maps the reference pressure ($u_r$) to the joint position ($v(r)$) for quasi-static pressure trajectories. Right: The empirical transfer function, based on spectral analysis, to characterize the pressurized air and soft coupling dynamics for the small and large pouch motors used in this work. (c) Representative results of the proposed inverse dynamic filter based on the identified plant dynamics from (b). The inverse dynamic filter was cascaded with a parameter-varying filter $(H(z,\rho))$ that attenuated high frequency dynamics based on the scheduling variable $\rho$}.
\label{fig4}
\vspace{-15pt}
\end{figure}

We identified our Hammerstein dynamic model, mapping the reference pressure to the joint angle, using 140~s of data from the user. To implement our controller, we derived an approximate inverse of each of the components in our model (see Fig.~\ref{fig4}(a)(ii)). Because the Everett contour can model both clockwise (e.g., forward) and counter-clockwise (e.g., inverse) hysteresis loops, estimating the inverse can be accomplished by simply swapping the input and output signals to generate a new surface for interpolation. 

Because direct inversion of the identified plant $P(z)$ can result in a filter that is both unstable and non-causal, we estimated an approximate inverse \cite{widrow1987adaptive}. We modeled the inverse as a second order discrete time transfer function.
\begin{equation}
\begin{aligned}
    F(z, \phi) = \frac{c_0 +c_1z^{-1} + c_2z^{-2}}{1+d_1z^{-1}+d_2z^{-2}} \\
     \phi = [c_0, ~c_1,~c_2,~d_1,~d_2]
\end{aligned}
\end{equation}
\noindent where $\phi$ contained the model parameters that defined the location of the poles and zeros. We observed that higher order filters led to a more complex controller with negligible improvements in the performance. Because the goal was only to capture the dynamic effects of the human-robot system, we constrained the optimization to enforce unity DC gain.
\begin{equation}
c_0 = 1 + d_1 + d_2 - c_1 - c_2
\end{equation}
This ensured that during quasi-static movements from the user, the output from the controller was only dependent on the nonlinear static map. We minimized the following two-norm to identify the optimal values for $F(z,\phi)$. 
\begin{equation}
    \hat{\phi}
    =
    \arg\min_{\phi_F}
    \left\|
    \Phi_L(M(e^{j\omega}) - F(e^{j\omega},\phi)P(e^{j\omega})) 
    \right\|_2^2
    \label{eq17}
\end{equation}
\noindent where $z=e^{jw}$. We introduced a 4th-order low pass Butterworth filter $(M(z))$ with a cutoff frequency of 10 Hz to prevent unbounded high frequency gains. We generated a synthetic dataset from the plant $P(z)$ seen in Eq.~\ref{eq15} using white noise with unit variance passed through a filter $(L(z))$ to shape the spectral energy.
\begin{equation}
    \Phi_{L} = |L(e^{j\omega})|^2
\end{equation}
We chose $L(z)$ as a 1st-order low pass filter with a cutoff frequency of 0.5 Hz to prioritize an accurate approximation of the inverse in the lower frequency band of interest for human motion. We minimized the error function shown in Eq.~\ref{eq17} using the same solver to identify $P(z, \lambda)$.

To attenuate high frequency noise during quasi-static motions while preserving the dynamics of our inverse-plant model during faster motions, we introduced a linear parameter varying (LPV) filter \cite{shamma2012overview} that was applied in series with $F(z,\phi)$ (see $H(z,\rho)$ in Fig.~\ref{fig4}(a)ii).

The parameter varying filter had the following structure.
\begin{equation}
    \hat\theta^+_f[t] = \gamma(1-\rho)\hat\theta^+_f[t-1] + (1-\gamma(1-\rho))\hat\theta^+[t]
    \label{eq19}
\end{equation}
\noindent where $\gamma$ was a factor that controlled how aggressive the filter attenuated high frequency signals and $\rho$ was a scheduling variable defined as follows.
\begin{equation}
\rho(\Delta \theta) =
\begin{cases}
0 & \Delta \theta < \Delta\theta_{LB} \\[3pt]
\frac{\Delta \theta - \Delta\theta_{LB}}{ \Delta\theta_{UB} - \Delta\theta_{LB}} & \Delta\theta_{LB} \le \Delta \theta \le \Delta\theta_{UB} \\[3pt]
1 & \Delta \theta > \Delta\theta_{UB}
\end{cases}
\label{eq20}
\end{equation}
$\Delta\theta$ was the difference between the measured position and the output from $F(z,\phi)$. $\Delta\theta_{LB}$ and $\Delta\theta_{UB}$ bounded the region where $\rho$ grew linearly from 0 to 1. For slow movements, $\Delta\theta \to 0$ and hence $\rho \to 0$ which caused Eq.~\ref{eq19} to function as a low pass filter with gain $\gamma$. For fast movements, $\Delta\theta$ increased, which caused $\rho \to 1$ and the parameter varying filter to function like a unity gain to preserve the dynamics of the $F(z, \phi)$. By incorporating a smooth transition using $\rho$, the controller can seamlessly transition between quasi-static and dynamic tasks when necessary. In this work, we empirically tuned the parameters based on preliminary experiments to achieve the best performance: $\alpha = 0.83$, and $[\Delta\theta_{LB}, \Delta\theta_{UB}] = [5\textdegree, 10\textdegree]$

With the inverse components of our Hammerstein dynamic model, we generated the controller using a Wiener dynamic structure \cite{schoukens2017identification} to integrate into a positive feedback loop (see Fig.~\ref{fig4}(a)ii) for human-in-the-loop assistance. 

\begin{figure*}[!t]
\centerline{\includegraphics[width=1\textwidth]{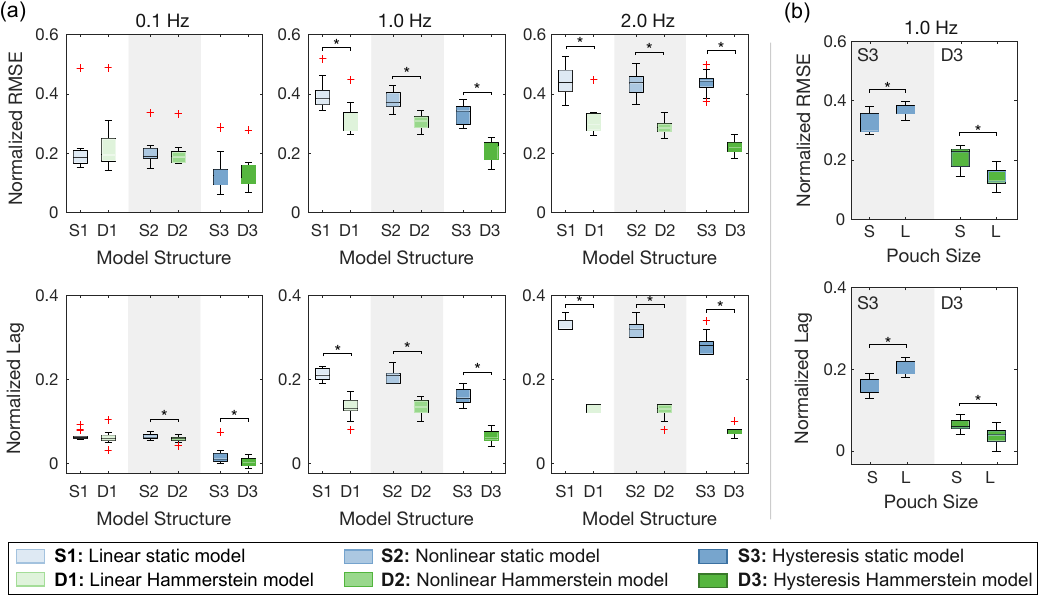}}
\vspace{-5pt}
\captionsetup{labelformat=empty}
\caption{Summary of the performance of different model structures when used as an inverse-plant feedforward controller. (a) Normalized root mean squared tracking error (NRMSE) and normalized lag between the reference and actual position for different sinusoidal trajectories when using the small pouch geometry. The dynamic models (D1-D3) significantly outperformed the static models (S1-S3) for the 1.0 Hz and 2.0 Hz trajectories. (b) Performance of the static (S3) and dynamic (D3) model for different pouch geometries. The dynamic model accounted for the changes in air flow dynamics between the small and large pouches to achieve high tracking accuracy regardless of the pouch geometry. Differences between groups that were statistically significant were marked by an asterisk.}
\label{fig5}
\vspace{-15pt}
\end{figure*}

\section{Inverse-Plant Model Evaluation}

We conducted a user study to evaluate the accuracy of our Hammerstein dynamic modeling approach when compared to state-of-the-art models that do not take into account dynamics.

\subsection{User Study Protocol}

We first collected the calibration data for each user to identify the parameters for the inverse-plant model. We informed each user to stay relaxed while we collected the calibration data.

We identified three static models (S1-S3) and three dynamic models (D1-D3). The static models were as follows: (S1) A linear model where we assumed that the joint could be modeled as a linear spring with an equilibrium position at 0 degrees, (S2) A nonlinear model where we assumed that the joint could be modeled as a 4th order polynomial with a non-zero equilibrium position, and (S3) the Preisach hysteresis model as identified using the method outlined in Section III. We augmented each of the three static models with a linear time invariant filter to generate the dynamic models (D1-D3) using the methods outlined in this work. For this study, we did not include the springs in the test rig to ensure that the models were only capturing the dynamics of the human and robot.

We used the six models as inverse-plant feedforward controllers to track predefined sinusoidal trajectories. We evaluated the accuracy of each model by calculating the root mean squared error (RMSE) and lag between the reference and measured trajectories. We calculated the cross correlation between the reference and measured trajectory with different delays and estimated the lag based on which delay had the largest cross correlation score. We normalized the lag and RMSE using the period and peak-to-peak amplitude of the reference trajectory.

We evaluated three reference trajectories with different frequencies (0.1 Hz, 1.0 Hz, 2.0 Hz) each with an amplitude of 20\textdegree. For the 0.1 Hz trajectories, we collected 30 seconds of data to ensure we measured three full cycles of motion. For the 1.0 Hz and 2.0 Hz trajectories, we collected 10 seconds of data. When testing each of the feedforward controllers, we informed the user to stay relaxed because each model was calibrated on the passive impedance of their wrist. We pseudo-randomized the order of the different models and frequencies to minimize any familiarization effects that could occur as participants became more accustomed to using the test rig.

To compare the performance of the different models, we first used the Shapiro-Wilk test to evaluate normality in our data. If the data followed a normal distribution, we conducted a paired T-test for each static and dynamic model pair (i.e., S1 and D1, S2 and D2, S3 and D3). If the data was not normally distributed, we used the Wilcoxon Signed-Rank Test. We used a significance threshold of 0.05 and corrected the P-values for multiple comparisons between the models for each reference trajectory using the false discovery rate and Benjamini-Hochberg procedure.

\subsection{Results}
We recruited 12 participants (eight male, four female, Age: 18-36, all right-handed) to evaluate our modeling strategy. All users gave informed consent and the experiment was approved by XXX, Institutional Review Board (Study \# XXX). We conducted all tests on the user's dominant hand and used the small pouch geometry outlined in Section IIIA. 

For the 0.1 Hz reference trajectory, the dynamic models (D1-D3) had similar performance to the static models (S1-S3) (see Fig.~\ref{fig5}a). Models S3 and D3 had the best performance in terms of the normalized root mean squared error (NRMSE) (0.13 compared to 0.20 on average for all other models) and normalized lag (0.01 compared to 0.06 on average for all other models). While there was a statistically significant reduction in lag between S2/D2 and S3/D3, the static models still had excellent performance with normalized lag below 0.07 and 0.02 respectively. For the 1.0 Hz reference trajectory, the static models had considerably larger tracking errors and lag (0.36 NRMSE and 0.19 normalized lag on average for S1-S3) compared to the 0.1 Hz trajectory. The dynamic models, on the other hand, had significant improvements compared  to the static equivalents in both NRMSE and normalized lag, with the D3 model performing the best (0.21 NRMSE and 0.06 normalized lag for D3). For the 2.0 Hz reference trajectory, we saw similar trends where tracking errors exceeded 0.4 NRMSE and 0.28 normalized lag for the static models. The dynamic models significantly improved the tracking performance compared to the static models, with D3 having the lowest tracking errors once again (0.22 NRMSE and 0.07 normalized lag).

To see how our modeling strategy adapted to soft exosuits with different air flow dynamics, we evaluated the large pouch geometry using the 1.0 Hz reference trajectory (see Fig.~\ref{fig5}b). We selected models S3 and D3 for this experiment because they were the most comprehensive static and dynamic models that we evaluated and could capture the most information about the system. For the large actuator, the static model had worse tracking performance (12\% increase in NRMSE and 31\% increase in normalized lag) compared to the small actuator. The dynamic model on the other hand, captured the slower dynamics of the robot and compensated for them accordingly, leading to a NRMSE of 0.14 and normalized lag of 0.06.

\subsection{Discussion}

We evaluated three common static modeling approaches and how augmenting them into a Hammerstein dynamic model improved their performance as a feedforward controller over different frequencies and speeds. For the 0.1 Hz reference trajectory, each static model had comparable performance to their dynamic counterparts. This was expected given that the 0.1~Hz trajectory was within the open-loop bandwidth of the robot. Our results were comparable to existing work that evaluated different static models at quasi-static speeds ($<3^\circ/\text{s}$) \cite{mccann2025body}. For the 1.0~Hz and 2.0~Hz trajectories, all static models had a NRMSE above 0.35 and normalized lag above 0.19. By augmenting the static models with a dynamic filter, however, the tracking performance significantly improved, therefore expanding the range of speeds and frequencies that the model can capture. While we had to collect more data to identify the dynamic behavior of our system compared to static approaches, our entire data collection procedure was efficient and only took 140~s. This is much faster than existing control strategies that relied on machine learning \cite{arnold2025personalized} or reinforcement learning \cite{yumbla2025personalized} which required up to 20 minutes of data collection or learning. 

Regardless of the type of static model we evaluated, we saw significant improvements in performance when incorporating the inverse dynamic filter. This suggests that, while improving the fidelity of the static model can improve the tracking accuracy (i.e., using a hysteresis model compared to a higher order polynomial), incorporating a component that captures dynamics can result in considerably larger improvements. Given that calibrating the hysteresis model required multiple concentric hysteresis loops that may need to be re-calibrated every time a user dons the device \cite{arnold2025personalized} and that a higher order polynomial can be calibrated using just the major hysteresis loop, there may be a modeling approach that balances model fidelity and calibration complexity for practical deployment. 

Pneumatic actuators with larger volumes can be more desirable for fully wearable systems because they can distribute forces over large regions on the body and require lower pressures during real-world deployment. Larger actuators, however, generally have lower bandwidths due to their higher flow requirements \cite{mosadegh2014pneumatic}. By incorporating an inverse dynamic filter into our model, any changes in the dynamics of the robot (e.g., the bandwidth) can be compensated for that would otherwise be ignored by a static model. To our surprise, the performance of the dynamic model with the large actuators was better than with the small actuators (see Fig.~\ref{fig5}(b)). We believe, however, this was because the slower air flow dynamics dominated other dynamic effects, such as the compliant coupling between the actuators and the test rig or the nonlinearities due to the pressure regulators that ultimately improved modeling accuracy. 

\section{Human-in-the-Loop Controller Evaluation}

We conducted a second user study to evaluate our inverse-plant model when integrated into the positive feedback loop with the parameter varying filter (see Fig.~\ref{fig4}(a)). We used the model D3 for our controller because it had the best performance over the frequencies and speeds of interest. We also evaluated model S3 to see how incorporating dynamics into the control loop could improve the quality of assistance. Lastly, we evaluated a no assistance case to function as the baseline condition. We referred to each of these controller conditions as no control (no assistance) Static (S3) and Dynamic (D3) for the remainder of this work. 

\subsection{User Study Protocol}

We collected the calibration data with the springs in the test rig to simulate the elevated stiffness in the wrist typically after a stroke (see Fig.~\ref{fig3}(c)). For this study, we monitored the electromyography signals (Trigno Lite, Delsys) of the flexor carpi ulnaris (FCU) and flexor carpi radialis (FCR) for wrist flexion. We also monitored the extensor capri ulnaris (ECU) and extensor carpi radialis (ECR) for wrist extension. We sampled the EMG signals at 1037 Hz, bandpass filtered the raw data between 20 and 450 Hz, and applied a 300 ms root mean square (RMS) sliding window. We normalized the signals using the max voluntary contraction of each user during isometric flexion and extension experiments before beginning the study. For all the experiments in this study, participants performed a marker following task where they flexed and extended their wrist to follow a reference trajectory displayed on a monitor. We gave the participants a brief training period to familiarize themselves with the emulator and task, during which they practiced the task with the no control condition.

We first evaluated our inverse-plant control strategy for different speeds and frequencies relevant for daily living activities. The reference trajectories were periodic triangular waveforms with an amplitude of 20\textdegree\ and speeds ranging from 8\textdegree/s (quasi-static motion) to 120\textdegree/s (highly dynamic motion), corresponding to frequencies between 0.1 Hz and 1.5 Hz. These speeds cover approximately 80\% of wrist motion during activities of daily living \cite{anderton2023movement}. For each trajectory, we evaluated three controller conditions: no control (no assistance), static control (S3), and dynamic control (D3). For the 8\textdegree/s trajectory, we also evaluated our dynamic control without the parameter varying filter (see $H(z,\rho)$ in Fig.~\ref{fig4}). We gave each user three trials for each control condition and started with the no control case for each trajectory before pseudo-randomizing the order of the static and dynamic control conditions. We evaluated the dynamic control without the parameter varying filter at the end of the 8\textdegree/s trajectory tests.

We conducted a final target reaching task by randomizing several positions (+/-20\textdegree, +/-10\textdegree\ and 0\textdegree) for the user to reach. Each position was repeated three times for a total of 15 positions. The users were told to move to each of the targets at a speed they were comfortable with. Once the user was within 2.5\textdegree\ of the target for 0.5 seconds, the target moved to the next position. We once again tested the no control condition first and gave each user three trials for each control condition. It should be noted that while we programmed the reference trajectories based on the goals of the study, the controllers only relied on kinematic information based on the user's voluntary motion to provide assistance.

We used a linear mixed effect model to determine whether the control condition had a significant effect on different metrics related to the user's performance (e.g., muscle activation, interaction torque, tracking accuracy). We selected our fixed effects as the control condition (i.e., No Control, Static, and Dynamic) and the trial number to account for any potential learning effects during the study. We set the random effect as the user to account for differences in baseline performance among participants.

We first fit a linear mixed-effects model that included the interaction effect between the controller condition and the trial number to determine whether learning effects differed across controllers. If the interaction was not statistically significant, we removed it from the model. We then identified a new model to determine if there was a common learning effect across all controller conditions. If trial number was not significant (i.e., participants showed no change in performance across the trials), it was also removed, leaving controller condition as the only fixed effect in our model. We used a significance threshold of 0.05 and corrected the P-values for multiple comparison between the control conditions for each reference trajectory using the false discovery rate and Benjamini-Hochberg procedure.

\begin{figure}[t]
\centerline{\includegraphics[width=0.49\textwidth]{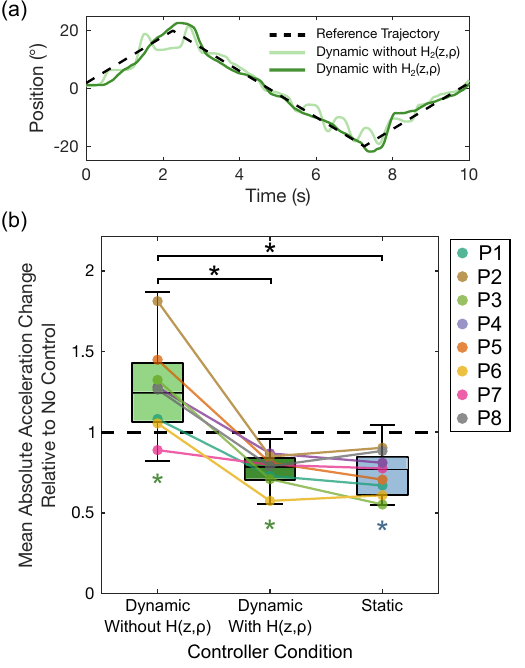}}
\vspace{-5pt}
\captionsetup{labelformat=empty}
\caption{Effects of the parameter varying filter, $H(z,\rho)$, on the mean absolute acceleration for the 8\textdegree/s (0.1 Hz) trajectory. (a) Representative results for the dynamic controller from a single trial. Without $H(z,\rho)$, the inverse dynamic controller amplified high frequency noise which led to non-smooth motions. (b) Mean absolute acceleration results for all users and test conditions. Statistically significant differences between controllers were marked with the black asterisks while statistically significant differences between the controllers and the no control condition were marked with the colored asterisks.}
\label{fig6}
\vspace{-15pt}
\end{figure}

\subsection{Results}

We recruited 8 participants (6 Male, 2 Female, Age: 19-30, 7 right handed, 1 left handed) for the second user study. All users gave informed consent and the experiment was approved by XXX, Institutional Review Board (Study \# XXX). We conducted all experiments on the user's dominant hand and used the pouch actuator with the small geometry.

\subsubsection{Parameter Varying Filter Performance}
Because the reference trajectory was a quasi-static triangular waveform, smoother motions were correlated with smaller accelerations. As a result, we estimated a smoothness score based on the mean absolute acceleration over the course of the trial to evaluate the quality of movement for the 0.1 Hz trajectory (see Fig.~\ref{fig6}). We normalized each of the controller conditions with the no control case to minimize any variance due to the baseline performance of the user. 

We observed a significant effect for the trial number on the mean absolute acceleration that suggested users improved their smoothness scores for successive trials ($-3.246^{\circ}/\text{s}^2$ per trial, $p=7.0\times10^{-3}$). There was, however, no significant interaction effect between the controllers. Without the parameter varying filter, $H(z,\rho)$, the inverse dynamic model amplified high frequency noise that led to oscillations and ultimately a 25\% increase $(p=3.7\times10^{-9})$ in the mean absolute accelerations compared to the no control condition (see Fig.~\ref{fig6}). When we incorporated the parameter varying filter, however, the mean absolute acceleration reduced by 24\% relative to the no control case $(p=1.0\times10^{-8})$. This was comparable to the static controller that did not consider any dynamics, which reduced the acceleration by 26\% compared to the no control case $(p=1.2\times10^{-9})$. There was no statistically significant difference between the dynamic controller (with $H(z,\rho)$) and the static controller.

\subsubsection{Inverse-plant Control for Different Speeds and Frequencies}

\begin{figure}[t]
\centerline{\includegraphics[width=0.5\textwidth]{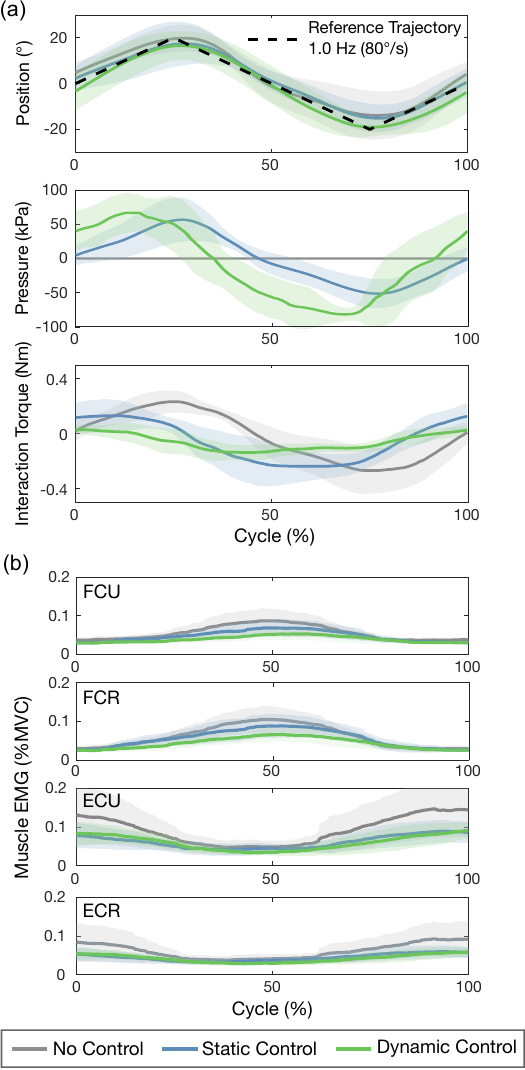}}
\vspace{-5pt}
\captionsetup{labelformat=empty}
\caption{Representative results from a user for the 80\textdegree/s (1.0 Hz) trajectory. (a) Participants tried to follow a reference trajectory that was presented on the screen. We recorded their position, the calculated reference pressure based on the selected controller, and the interaction torque between the user and the robot. (b) We also monitored the activation of the flexor carpi ulnaris (FCU), flexor carpi radialis (FCR), extensor carpi ulnaris (ECU), and extensor carpi radialis (ECR) muscles to evaluate the physiological effects of the different controllers.}
\label{fig7example}
\vspace{-15pt}
\end{figure}

We evaluated our control strategy for different speeds and frequencies by measuring the interaction force between the user and the test rig, electromyography signals from the muscles responsible for flexing and extending the wrist, and tracking accuracy with respect to the reference trajectory (see Fig.~\ref{fig7example} for a representative example). For all analysis on the interaction torque and muscle activation, there was no statistically significant learning effects based on the trial number.

We evaluated the net change in the interaction torque relative to the no control condition for the different speeds and frequencies tested. For the 0.1 Hz trajectory, the static and dynamic controllers had similar performance and reduced the interaction torque by 64\% $(p=4.2\times10^{-28})$ and 73\% $(p=4.3\times10^{-30})$. For the 0.5 Hz trajectory, the static and dynamic controllers had similar performance once again and reduced the interaction torque by 48\% $(p=4.9\times10^{-20})$ and 50\% $(p=1.3\times10^{-20})$. For both of these trajectories, there was no statistically significant difference between the static and dynamic controllers. For the faster trajectories, the dynamic controller significantly outperformed the static controller. The dynamic controller reduced the interaction torque by 40\% $(p=3.8\times10^{-28})$ for the 1.0 Hz trajectory and by 28\% $(p=3.8\times10^{-14})$ for the 1.5 Hz trajectory. The static controller reduced the interaction torque by only 8\% $(p=2.5\times10^{-04})$ and increased the interaction torque by 12\% $(p=2.8\times10^{-6})$ for the 1.0 Hz and 1.5 Hz trajectory, respectively.

We observed similar changes in the mean muscle activation that we observed from the interaction torque for the static and dynamic controllers relative to the no-control condition. For the 8\textdegree/s (0.1 Hz) trajectory, there was a 51\% reduction in activation of the flexor carpi ulnaris (FCU, $p=3.4\times10^{-12}$) and flexor carpi radialis (FCR, $p=2.37\times10^{-15}$) for the static controller. There was a 47\% reduction in activation of the FCU $(p=4.4\times10^{-11})$ and FCR $(p=4.1\times10^{-14})$ for the dynamic controller. We observed a 25\% $(p=1.5\times10^{-3})$ and 40\% $(p=7.2\times10^{-14})$ reduction in the activation of the extensor carpi ulnaris (ECU) and extensor carpi radialis (ECR) for the static controller, and a 30\% reduction $(p=3.0\times10^{-4})$ and 38\% reduction $(p=1.8\times10^{-13})$ in activation of the ECU and ECR for the dynamic controller. There was no statistically significant difference between the static and dynamic controller for any of the mean muscle activations.

\begin{figure*}[!t]
\centerline{\includegraphics[width=1\textwidth]{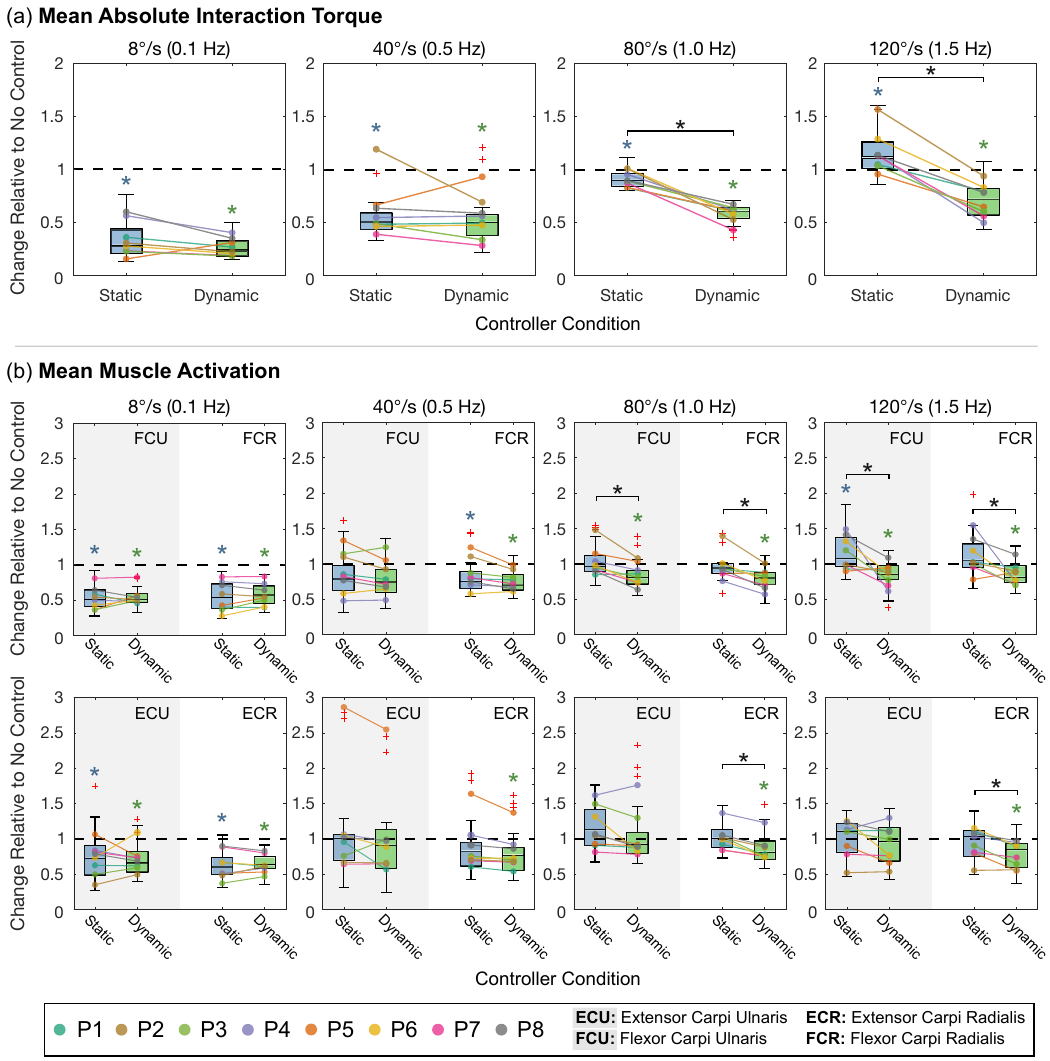}}
\vspace{-5pt}
\captionsetup{labelformat=empty}
\caption{Interaction torque and muscle activation results for trajectories with reference speeds ranging from 8\textdegree/s (0.1 Hz) to 120\textdegree/s (1.5 Hz). (a) Change in the mean absolute interaction torque for the different control conditions. The dynamic controller outperformed the static controller for the 80\textdegree/s (1.0 Hz) and 120\textdegree/s (1.5 Hz) trajectories. (b) Change in the mean muscle activation for the flexor and extensor muscles in the wrist. The dynamic controller once again provided more effective assistance for the 80\textdegree/s (1.0 Hz) and 120\textdegree/s (1.5 Hz) trajectories. Statistically significant differences between controllers were marked with the black asterisks while statistically significant differences between the controllers and the no control case were marked with the colored asterisks.}
\label{fig7}
\vspace{-15pt}
\end{figure*}

For the 40\textdegree/s (0.5 Hz) trajectory, we saw no statistically significant change in activation of the FCU and ECU for the static and dynamic controllers. We observed, however, a 23\% reduction $(p=8.36\times10^{-6})$ and 28\% reduction $(p=1.7e^{-7})$ in muscle activation for the FCR for the static and dynamic controllers. The static controller had no statistically significant effect on the activation of the ECR, however, the dynamic controller had an 18\% reduction $(p=0.027)$ in activation of the ECR.

For the 80\textdegree/s (1.0 Hz) trajectory, we began to observe significant differences in performance between the static and dynamic controllers. Relative to the no control condition, we observed changes up to 10\% (in both increasing and decreasing directions) of the flexor and extensor muscles for the static controller, however none of which were statistically significant. For the dynamic controller, we observed a 15\% reduction $(p=1.2\times10^{-3})$ and 21\% reduction $(p=6.0\times10^{-10})$ in activation of the FCU and FCR, as well as a 17\% reduction $(p=1.1\times10^{-3})$ in the ECR. We observed no significant change in the activation of the ECU.

For the 120\textdegree/s (1.5 Hz) trajectory, we continued to observe significant differences in performance between the static and dynamic controllers. The static controller had no statistically significant effect on the FCR, ECU, and ECR, relative to the no control condition, however, there was a 10\% increase $(p=0.032)$ in activation from the FCU. The dynamic controller had a 13\% reduction on the activation of the FCU $(p=0.01)$ and FCR $(p=3.0\times10^{-3})$ and a 26\% reduction on the activation for ECR $(p=1.47\times10^{-5})$. While our dynamic controller had a 14\% reduction in activation of the ECU, the results were not statistically significant.

We also evaluated the tracking accuracy of each user to see how each controller influenced the participants' ability to complete the marker following task. For the 40\textdegree/s (0.5 Hz) trajectory, we observed a statistically significant effect of the trial number on the tracking accuracy of the user (tracking RMSE changed by -0.448\textdegree/trial, $p=0.025$), however, for all other conditions there was no significant effect. The static controller had a negative effect on the tracking accuracy for the 0.1 Hz trajectory $(p=0.015)$ and 1.0 Hz trajectory $(p=0.033)$, however, the effect was small ($<1^{\circ}$ increase in tracking error). All other control conditions showed no effect on tracking performance relative to the no control condition.

\subsubsection{Inverse-Plant Control for Target Reaching Tasks}
\begin{figure}[t]
\centerline{\includegraphics[width=0.49\textwidth]{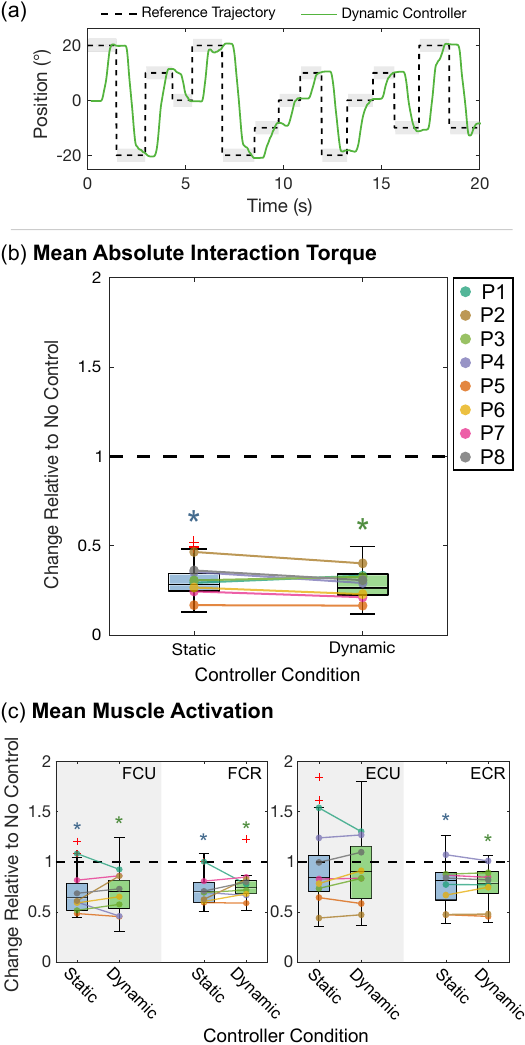}}
\vspace{-5pt}
\captionsetup{labelformat=empty}
\caption{Interaction torque and muscle activation results for the target reaching task. (a) Representative results from a user for the target reaching task when testing the dynamic controller. (b) Change in the mean absolute interaction torque between the user and the test rig for the static and dynamic controller conditions. (c) Change in the mean muscle activation for the flexor and extensor muscles in the wrist. We observed similar performance between the static and dynamic controllers relative to the no control condition. Statistically significant differences between the controllers and the no control case were marked with the colored asterisks.}
\label{fig8}
\vspace{-15pt}
\end{figure}

We once again evaluated the interaction torque and mean muscle activation of the users for the target reaching task. There was no significant learning effects based on the trial number for the interaction torque or muscle activation. Both the static and dynamic controllers significantly reduced the interaction torque between the user and the test rig by approximately 70\% (static controller: $p=1.96\times10^{-40}$, dynamic controller: $p=5.04\times10^{-41})$ relative to the no control condition (see Fig.~\ref{fig8}(a)). The static and dynamic controllers also reduced the activation of all evaluated muscles by similar amounts relative to the no control condition. The activation of the FCU reduced by 21\% ($p=6.56\times10^{-9}$ for both the static and dynamic controllers) and the activation of the FCR reduced by 27\% (static controller: $p=5.58\times10^{-10}$, dynamic controller: $p=8.06\times10^{-10}$). The activation of the ECR reduced by 30\% for both controllers ($p=9.55\times10^{-6}$ for both the static and dynamic controllers). The activation of the ECU reduced by approximately 14\% for both the static and dynamic controllers, however there was no statistical significance.

We also evaluated the tracking accuracy and completion time and found that the trial number had a significant effect on the finish time ($-0.863s/trial$, $p=0.032$) but no interaction effect with the control conditions. There was no significant effect for the trial number on the tracking accuracy. There was also no statistically significant difference for the tracking accuracy or the finish time between any of the controllers and no control condition.

\subsection{Discussion}

Incorporating the parameter varying filter into the feedback loop prevented noise amplification when the user was moving slowly, resulting in similar smoothness scores as the static controller. As the motion became faster and more dynamic, however, the parameter varying filter simplified to a unity gain, preserving the dynamic properties of the controller. This is essential for daily living activities which require assistance over a broad range of speeds and scenarios. Interestingly, there was a $\sim25\%$ reduction on the mean absolute acceleration compared to the no control condition when using the static and dynamic controllers with the parameter varying filter (see Fig.~\ref{fig6}). We believed this was because the users had to overcome the stiffness of the springs in the emulator during the no control condition. With the static and dynamic controllers, however, users felt minimal impedance from the springs, therefore making it easier to achieve a smooth trajectory. 

Our dynamic controller provided significant reductions in user effort, measured both by the interaction torque between the user and the emulator as well as the muscle activation of key flexor and extensor muscles for speeds up to 120\textdegree/s (1.5 Hz) relative to the no control condition. By incorporating dynamics into the controller, the calculated reference pressures have phase lead relative to the measured position that increased as the participants moved faster to compensate for the air flow and coupling dynamics (see Fig.~\ref{fig7example}(a) for representative example). The static controller, which only compensated for hysteresis and stiffness in the joint, generated a reference pressure that was in phase with the measured position (see Fig.~\ref{fig7example}(a)). As a result, the static controller only provided effective assistance up to 40\textdegree/s (0.5~Hz). For the 120\textdegree/s (1.5 Hz) trajectory, the assistance was out of the phase with the user, ultimately leading to an increase in interaction torque and muscle activation in the FCU. Based on existing literature that evaluated the dynamics of wrist motion during daily living activities, approximately 80\% of wrist motion occurs at speeds up to 120\textdegree/s/\cite{anderton2023movement} and 5 Hz covers approximately 75\% of the spectral density \cite{mann1989frequency}. Our inverse-plant control strategy, when incorporating dynamics into the model, can provide more effective assistance over these speeds and frequencies than existing strategies that do not take into account dynamics.

For the target reaching task, we observed similar performance between the static and dynamic controllers, with both improving tracking accuracy and reducing interaction torque and muscle activation relative to the no control condition. For the single frequency trajectories, we evaluated how the static and dynamic controllers performed for continuous motions at specific speeds of interest. For the target reaching task, we evaluated stop and start movements where the speed was self-selected by the users. We determined that the average speed that the users selected for each target reaching trial was $15.32 \pm 2.18^\circ/s$. This speed was within the range where both the static and dynamic controllers had similar performance for the single frequency trajectories, hence the results were as expected.


Our human-in-the-loop control strategy using our inverse-plant method provided two benefits compared to the state of the art: (1) it compensated for the slow dynamics seen in pneumatic actuators due to inherent air flow dynamics, and (2) it eliminated the need to empirically tune a dynamic model of the limb \cite{xiloyannis2019physiological}, or low-level pressure controller \cite{arnold2025personalized, realmuto2022assisting} that has an influence on the responsiveness and quality of assistance from the robot. Our control strategy also does not rely on physiological sensing \cite{lotti2020adaptive} or interaction force sensing \cite{arnold2025personalized, xiloyannis2019physiological, chiaradia2021assistive}, therefore reducing the number of components necessary for real-world deployment.

Our controller can be applied to any soft assistive exosuit with similar underlying properties to improve the range of speeds and frequencies over which the device can provide assistance. Based on our work, this is most suitable for exosuits that use textile materials and pneumatic actuators due to the hysteretic properties of the constituent materials and air flow dynamics. While we observed that larger-volume actuators had lower bandwidths, we selected the small pouch geometry when evaluating our control strategy to maximize the open-loop bandwidth of the robot and, consequently, the performance of our static controller. For other devices, such as soft exosuits for the shoulder \cite{arnold2025personalized}, elbow \cite{nassour2021soft}, or knee \cite{yumbla2025personalized}, which leveraged larger-volume actuators, we expect our proposed control strategy to provide an even greater performance benefit relative to a static controller.

\section{Limitations and Future Work}

In this work, we validated our control strategy on able-bodied users and on an exosuit emulator in a virtual marker following task. Future work should transfer the controller to a fully wearable soft exosuit and evaluate the controller in real world scenarios and populations that are clinically relevant. This would require addressing challenges such as wearable sensing and efficient coupling of the actuators to the body. Existing work has demonstrated an easy-to-fabricate sensing garment for continuous kinematic sensing that can be integrated into existing soft assistive exosuits \cite{sepehri2025retrofitting}. Other work has also shown a hybrid rigid-soft coupling mechanism to efficiently transfer loads to the body for assistive devices \cite{chiaradia2021assistive}.

We focused on a model-based approach to capture the nonlinear stiffness, hysteresis, and dynamic behavior seen in soft assistive exosuits. Future work should investigate augmenting our approach with model-free or learning-based techniques to address challenges such as: distinguishing between voluntary and involuntary movements, and mitigating abnormal synergies across multiple joints. While these effects are not a concern for human augmentation in individuals without motor impairments, they can have a substantial impact for applications in rehabilitation where the interaction between the user and the exosuit can be influenced by motor impairments due to a neurological injury.


\section{Conclusion}
In this work, we presented an inverse-plant control strategy for soft assistive exosuits that accounted for nonlinear stiffness, hysteresis, and human-robot dynamics to provide continuous assistance over the speeds and frequencies necessary for daily living activities. Our control method required only 140~s of data to personalize the controller to each user and inferred user intention solely from kinematic sensing to provide task-agnostic assistance. We evaluated our control strategy on a test rig that emulated a soft assistive exosuit for the wrist and observed up to 73\% reductions in interaction torque and up to 47\% reduction in activation of key flexor and extensor muscles in the wrist relative to the no control condition. Overall, our work demonstrated that explicitly accounting for the dynamic behavior when wearing a soft assistive exosuit can result in more effective assistance across quasi-static and dynamic movements without the need for force or physiological sensing during deployment.

\bibliographystyle{IEEEtran}
\bibliography{biblio}

@inproceedings{kazerooni2005control,
  title={On the control of the berkeley lower extremity exoskeleton (BLEEX)},
  author={Kazerooni, Hami and Racine, J-L and Huang, Lihua and Steger, Ryan},
  booktitle={Proceedings of the 2005 IEEE international conference on robotics and automation},
  pages={4353--4360},
  year={2005},
  organization={IEEE}
}

@article{kawato1999internal,
  title={Internal models for motor control and trajectory planning},
  author={Kawato, Mitsuo},
  journal={Current opinion in neurobiology},
  volume={9},
  number={6},
  pages={718--727},
  year={1999},
  publisher={Elsevier}
}

@article{charles2011dynamics,
  title={Dynamics of wrist rotations},
  author={Charles, Steven K and Hogan, Neville},
  journal={Journal of biomechanics},
  volume={44},
  number={4},
  pages={614--621},
  year={2011},
  publisher={Elsevier}
}

@book{Ljung,
  title={System Identification: Theory for the User},
  author={Ljung, Lennart},
  year={1999},
  publisher={Prentice Hall}
}

@article{schoukens2017identification,
  title={Identification of block-oriented nonlinear systems starting from linear approximations: A survey},
  author={Schoukens, Maarten and Tiels, Koen},
  journal={Automatica},
  volume={85},
  pages={272--292},
  year={2017},
  publisher={Elsevier}
}

@article{mayergoyz1986mathematical,
  title={Mathematical models of hysteresis},
  author={Mayergoyz, Isaac},
  journal={IEEE Transactions on magnetics},
  volume={22},
  number={5},
  pages={603--608},
  year={1986},
  publisher={IEEE}
}

@article{winters1988analysis,
  title={An analysis of the sources of musculoskeletal system impedance},
  author={Winters, Jack and Stark, Lawrence and Seif-Naraghi, Amir-Hussein},
  journal={Journal of biomechanics},
  volume={21},
  number={12},
  pages={1011--1025},
  year={1988},
  publisher={Elsevier}
}

@article{mosadegh2014pneumatic,
  title={Pneumatic networks for soft robotics that actuate rapidly},
  author={Mosadegh, Bobak and Polygerinos, Panagiotis and Keplinger, Christoph and Wennstedt, Sophia and Shepherd, Robert F and Gupta, Unmukt and Shim, Jongmin and Bertoldi, Katia and Walsh, Conor J and Whitesides, George M},
  journal={Advanced functional materials},
  volume={24},
  number={15},
  pages={2163--2170},
  year={2014},
  publisher={Wiley Online Library}
}

@article{proietti2021sensing,
  title={Sensing and control of a multi-joint soft wearable robot for upper-limb assistance and rehabilitation},
  author={Proietti, Tommaso and O’Neill, Ciar{\'a}n and Hohimer, Cameron J and Nuckols, Kristin and Clarke, Megan E and Zhou, Yu Meng and Lin, David J and Walsh, Conor J},
  journal={IEEE Robotics and Automation Letters},
  volume={6},
  number={2},
  pages={2381--2388},
  year={2021},
  publisher={IEEE}
}

@article{lotti2022myoelectric,
  title={Myoelectric or force control? A comparative study on a soft arm exosuit},
  author={Lotti, Nicola and Xiloyannis, Michele and Missiroli, Francesco and Bokranz, Casimir and Chiaradia, Domenico and Frisoli, Antonio and Riener, Robert and Masia, Lorenzo},
  journal={IEEE Transactions on Robotics},
  volume={38},
  number={3},
  pages={1363--1379},
  year={2022},
  publisher={IEEE}
}

@article{heisser2026codevelopment,
  title={The codevelopment of soft robotics and assistive technology},
  author={Heisser, Ronald H and Raman, Ritu and Shepherd, Robert F},
  journal={Science Robotics},
  volume={11},
  number={111},
  pages={eaee0269},
  year={2026},
  publisher={American Association for the Advancement of Science}
}

@article{shamma2012overview,
  title={An overview of LPV systems},
  author={Shamma, Jeff S},
  journal={Control of linear parameter varying systems with applications},
  pages={3--26},
  year={2012},
  publisher={Springer}
}

@article{ma2014incidence,
  title={Incidence, prevalence, costs, and impact on disability of common conditions requiring rehabilitation in the United States: stroke, spinal cord injury, traumatic brain injury, multiple sclerosis, osteoarthritis, rheumatoid arthritis, limb loss, and back pain},
  author={Ma, Vincent Y and Chan, Leighton and Carruthers, Kadir J},
  journal={Archives of physical medicine and rehabilitation},
  volume={95},
  number={5},
  pages={986--995},
  year={2014},
  publisher={Elsevier}
}

@article{andrews1979sroke,
  title={Sroke recovery: He can but does he?},
  author={Andrews, Keith and Steward, Jean},
  journal={Rheumatology},
  volume={18},
  number={1},
  pages={43--48},
  year={1979},
  publisher={Oxford University Press}
}

@article{chen2026call,
  title={A call for a performance-driven approach for soft robotics research},
  author={Chen, Yufeng and Chirarattananon, Pakpong and Jayaram, Kaushik},
  journal={npj Robotics},
  volume={4},
  number={1},
  pages={14},
  year={2026},
  publisher={Nature Publishing Group UK London}
}

@article{schaffer2024soft,
  title={Soft wrist exosuit actuated by fabric pneumatic artificial muscles},
  author={Sch{\"a}ffer, Katalin and Ozkan-Aydin, Yasemin and Coad, Margaret M},
  journal={IEEE Transactions on Medical Robotics and Bionics},
  volume={6},
  number={2},
  pages={718--732},
  year={2024},
  publisher={IEEE}
}

@article{wolpert1998multiple,
  title={Multiple paired forward and inverse models for motor control},
  author={Wolpert, Daniel M and Kawato, Mitsuo},
  journal={Neural networks},
  volume={11},
  number={7-8},
  pages={1317--1329},
  year={1998},
  publisher={Elsevier}
}

@article{gorgey2018robotic,
  title={Robotic exoskeletons: The current pros and cons},
  author={Gorgey, Ashraf S},
  journal={World journal of orthopedics},
  volume={9},
  number={9},
  pages={112},
  year={2018}
}

@article{zhou2024portable,
  title={A portable inflatable soft wearable robot to assist the shoulder during industrial work},
  author={Zhou, Yu Meng and Hohimer, Cameron J and Young, Harrison T and McCann, Connor M and Pont-Esteban, David and Civici, Umut S and Jin, Yichu and Murphy, Patrick and Wagner, Diana and Cole, Tazzy and others},
  journal={Science Robotics},
  volume={9},
  number={91},
  pages={eadi2377},
  year={2024},
  publisher={American Association for the Advancement of Science}
}

@article{devittori2024unsupervised,
  title={Unsupervised robot-assisted rehabilitation after stroke: feasibility, effect on therapy dose, and user experience},
  author={Devittori, Giada and Dinacci, Daria and Romiti, Davide and Califfi, Antonella and Petrillo, Claudio and Rossi, Paolo and Ranzani, Raffaele and Gassert, Roger and Lambercy, Olivier},
  journal={Journal of neuroengineering and rehabilitation},
  volume={21},
  number={1},
  pages={52},
  year={2024},
  publisher={Springer}
}

@article{ochieze2023wearable,
  title={Wearable upper limb robotics for pervasive health: A review},
  author={Ochieze, Chukwuemeka and Zare, Soroush and Sun, Ye},
  journal={Progress in Biomedical Engineering},
  volume={5},
  number={3},
  pages={032003},
  year={2023},
  publisher={IOP Publishing}
}

@article{siviy2023opportunities,
  title={Opportunities and challenges in the development of exoskeletons for locomotor assistance},
  author={Siviy, Christopher and Baker, Lauren M and Quinlivan, Brendan T and Porciuncula, Franchino and Swaminathan, Krithika and Awad, Louis N and Walsh, Conor J},
  journal={Nature biomedical engineering},
  volume={7},
  number={4},
  pages={456--472},
  year={2023},
  publisher={Nature Publishing Group UK London}
}

@article{chen2024systematic,
  title={A systematic review on rigid exoskeleton robot design for wearing comfort: Joint self-alignment, attachment interface, and structure customization},
  author={Chen, Longbao and Zhou, Ding and Leng, Yuquan},
  journal={IEEE Transactions on Neural Systems and Rehabilitation Engineering},
  volume={32},
  pages={3815--3827},
  year={2024},
  publisher={IEEE}
}

@article{li2021design,
  title={Design and validation of a cable-driven asymmetric back exosuit},
  author={Li, Jared M and Molinaro, Dean D and King, Andrew S and Mazumdar, Anirban and Young, Aaron J},
  journal={IEEE Transactions on Robotics},
  volume={38},
  number={3},
  pages={1489--1502},
  year={2021},
  publisher={IEEE}
}

@article{kim2025exo,
  title={Exo-glove poly III: Grasp assistance by modulating thumb and finger motion sequence with a single actuator},
  author={Kim, Kyu Bum and Choi, Hyungmin and Kim, Byungchul and Kang, Brian Byunghyun and Cheon, Sangheui and Cho, Kyu-Jin},
  journal={Soft Robotics},
  volume={12},
  number={5},
  pages={593--605},
  year={2025},
  publisher={SAGE Publications Sage CA: Los Angeles, CA}
}

@article{yang2022soft,
  title={A soft exosuit assisting hip abduction for knee adduction moment reduction during walking},
  author={Yang, Hee Doo and Cooper, Myles and Eckert-Erdheim, Asa and Orzel, Dorothy and Walsh, Conor J},
  journal={IEEE Robotics and Automation Letters},
  volume={7},
  number={3},
  pages={7439--7446},
  year={2022},
  publisher={IEEE}
}

@article{krebs1998robot,
  title={Robot-aided neurorehabilitation},
  author={Krebs, H Igo and Hogan, Neville and Aisen, Mindy L and Volpe, Bruce T},
  journal={IEEE transactions on rehabilitation engineering},
  volume={6},
  number={1},
  pages={75--87},
  year={1998},
  publisher={IEEE}
}

@article{perry2007upper,
  title={Upper-limb powered exoskeleton design},
  author={Perry, Joel C and Rosen, Jacob and Burns, Stephen},
  journal={IEEE/ASME transactions on mechatronics},
  volume={12},
  number={4},
  pages={408--417},
  year={2007},
  publisher={IEEE}
}

@article{zoss2006biomechanical,
  title={Biomechanical design of the Berkeley lower extremity exoskeleton (BLEEX)},
  author={Zoss, Adam B and Kazerooni, Hami and Chu, Andrew},
  journal={IEEE/ASME Transactions on mechatronics},
  volume={11},
  number={2},
  pages={128--138},
  year={2006},
  publisher={IEEE}
}

@inproceedings{sepehri2024soft,
  title={A soft robotic wrist orthosis using textile pneumatic actuators for passive rehabilitation},
  author={Sepehri, Anoush and Ward, Samuel and Tolley, Michael T and Morimoto, Tania K},
  booktitle={2024 IEEE 7th International Conference on Soft Robotics (RoboSoft)},
  pages={284--290},
  year={2024},
  organization={IEEE}
}

@article{peng2024improving,
  title={Improving complex task performance in powered upper limb exoskeletons with adaptive proportional myoelectric control for user motor strategy tracking},
  author={Peng, Xiangyu and Li, Shunzhang and Stirling, Leia},
  journal={IEEE Robotics and Automation Letters},
  volume={9},
  number={5},
  pages={4655--4662},
  year={2024},
  publisher={IEEE}
}

@article{nassour2021soft,
  title={Soft pneumatic elbow exoskeleton reduces the muscle activity, metabolic cost and fatigue during holding and carrying of loads},
  author={Nassour, John and Zhao, Guoping and Grimmer, Martin},
  journal={Scientific Reports},
  volume={11},
  number={1},
  pages={12556},
  year={2021},
  publisher={Nature Publishing Group UK London}
}

@inproceedings{sepehri2025retrofitting,
  title={Retrofitting Soft Assistive Robots with Sew-Free Sensing Garments for Joint Motion Tracking and Kinematic Feedback},
  author={Sepehri, Anoush and Tolley, Michael T and Morimoto, Tania K},
  booktitle={2025 International Conference On Rehabilitation Robotics (ICORR)},
  pages={1--7},
  year={2025},
  organization={IEEE}
}

@article{ferdousi2025complicacy,
  title={Complicacy in electrode position shift and its solution in sEMG pattern recognition: a review},
  author={Ferdousi, Arifa and Islam, Md Johirul and Ahmad, Shamim and Islam, Md Rezaul and Chowdhury, Md Nakib Hayat and Haque, Fahmida and Ali, Sawal Hamid Md and Reaz, Mamun Bin Ibne},
  journal={IEEE Sensors Journal},
  year={2025},
  publisher={IEEE}
}

@inproceedings{yumbla2025personalized,
  title={Personalized Reinforcement Learning Control of Soft Robotic Exosuit for Assisting Human Normative Walking with Reduced Effort},
  author={Yumbla, Emiliano Quinones and Zhong, Junmin and Soltanian, Seyed Yousef and Si, Jennie and Zhang, Wenlong},
  booktitle={2025 IEEE/RSJ International Conference on Intelligent Robots and Systems (IROS)},
  pages={18248--18248},
  year={2025},
  organization={IEEE}
}

@article{mann1989frequency,
  title={Frequency spectrum analysis of wrist motion for activities of daily living},
  author={Mann, Kenneth A and Wernere, Frederick W and Palmer, Andrew K},
  journal={Journal of Orthopaedic research},
  volume={7},
  number={2},
  pages={304--306},
  year={1989},
  publisher={Wiley Online Library}
}

@article{selen2006impedance,
  title={Impedance is modulated to meet accuracy demands during goal-directed arm movements},
  author={Selen, Luc PJ and Beek, Peter J and Van Die{\"e}n, Jaap H},
  journal={Experimental Brain Research},
  volume={172},
  number={1},
  pages={129--138},
  year={2006},
  publisher={Springer}
}

@article{crevecoeur2013priors,
  title={Priors engaged in long-latency responses to mechanical perturbations suggest a rapid update in state estimation},
  author={Crevecoeur, Fr{\'e}d{\'e}ric and Scott, Stephen H},
  journal={PLoS computational biology},
  volume={9},
  number={8},
  pages={e1003177},
  year={2013},
  publisher={Public Library of Science San Francisco, USA}
}

@article{niiyama2015pouch,
  title={Pouch motors: Printable soft actuators integrated with computational design},
  author={Niiyama, Ryuma and Sun, Xu and Sung, Cynthia and An, Byoungkwon and Rus, Daniela and Kim, Sangbae},
  journal={Soft Robotics},
  volume={2},
  number={2},
  pages={59--70},
  year={2015},
  publisher={Mary Ann Liebert, Inc. 140 Huguenot Street, 3rd Floor New Rochelle, NY 10801 USA}
}

@article{mccann2025body,
  title={On-body textile hysteresis estimation for personalized physical human-robot interaction},
  author={McCann, Connor M and Arnold, James and Lehmacher, Carolin and Bertoldi, Katia and Walsh, Conor J},
  journal={The International Journal of Robotics Research},
  pages={02783649251358840},
  year={2025},
  publisher={SAGE Publications Sage UK: London, England}
}

@article{park2014design,
  title={Design and control of a bio-inspired soft wearable robotic device for ankle--foot rehabilitation},
  author={Park, Yong-Lae and Chen, Bor-rong and P{\'e}rez-Arancibia, N{\'e}stor O and Young, Diana and Stirling, Leia and Wood, Robert J and Goldfield, Eugene C and Nagpal, Radhika},
  journal={Bioinspiration \& biomimetics},
  volume={9},
  number={1},
  pages={016007},
  year={2014},
  publisher={IOP Publishing}
}

@article{anderton2023movement,
  title={Movement preferences of the wrist and forearm during activities of daily living},
  author={Anderton, Will and Tew, Scott and Ferguson, Spencer and Hernandez, Joshua},
  journal={Journal of Hand Therapy},
  volume={36},
  number={3},
  pages={580--592},
  year={2023},
  publisher={Elsevier}
}

@article{wang2017neural,
  title={Neural and non-neural related properties in the spastic wrist flexors: an optimization study},
  author={Wang, Ruoli and Herman, Pawel and Ekeberg, {\"O}rjan and G{\"a}verth, Johan and Fagergren, Anders and Forssberg, Hans},
  journal={Medical Engineering \& Physics},
  volume={47},
  pages={198--209},
  year={2017},
  publisher={Elsevier}
}

@article{arnold2025personalized,
  title={Personalized ML-based wearable robot control improves impaired arm function},
  author={Arnold, James and Pathak, Prabhat and Jin, Yichu and Pont-Esteban, David and McCann, Connor M and Lehmacher, Carolin and Bonadonna, John P and Lewko, Tanguy and Burke, Katherine M and Cavanagh, Sarah and others},
  journal={Nature Communications},
  volume={16},
  number={1},
  pages={7091},
  year={2025},
  publisher={Nature Publishing Group UK London}
}

@article{daniels2023everett,
  title={Everett map construction for modeling static hysteresis: Delaunay-based interpolant versus B-spline surface},
  author={Daniels, Bram and Overboom, Timo and Curti, Mitrofan and Lomonova, Elena},
  journal={IEEE Transactions on Magnetics},
  volume={59},
  number={5},
  pages={1--4},
  year={2023},
  publisher={IEEE}
}

@article{zhang2017modeling,
  title={Modeling and inverse compensation of hysteresis in supercoiled polymer artificial muscles},
  author={Zhang, Jun and Iyer, Kaushik and Simeonov, Anthony and Yip, Michael C},
  journal={IEEE Robotics and Automation Letters},
  volume={2},
  number={2},
  pages={773--780},
  year={2017},
  publisher={IEEE}
}

@article{caputo2014universal,
  title={A universal ankle--foot prosthesis emulator for human locomotion experiments},
  author={Caputo, Joshua M and Collins, Steven H},
  journal={Journal of biomechanical engineering},
  volume={136},
  number={3},
  pages={035002},
  year={2014},
  publisher={American Society of Mechanical Engineers}
}

@article{shomron2025robotic,
  title={A robotic and virtual testing platform highlighting the promise of soft wearable actuators for wrist tremor suppression},
  author={Shomron, Alona Shagan and Chase-Markopoulou, Christina and Walter, Johannes R and Sellhorn-Timm, Johanna and Shao, Yitian and Nadler, Tobias and Benson, Audrey and Wochner, Isabell and Rumley, Ellen H and Wurster, Isabel and others},
  journal={Device},
  year={2025},
  publisher={Elsevier}
}

@article{bryan2021hip,
  title={A hip--knee--ankle exoskeleton emulator for studying gait assistance},
  author={Bryan, Gwendolyn M and Franks, Patrick W and Klein, Stefan C and Peuchen, Robert J and Collins, Steven H},
  journal={The International Journal of Robotics Research},
  volume={40},
  number={4-5},
  pages={722--746},
  year={2021},
  publisher={SAGE Publications Sage UK: London, England}
}

@incollection{widrow1987adaptive,
  title={Adaptive inverse control},
  author={Widrow, Bernard},
  booktitle={Adaptive Systems in Control and Signal Processing 1986},
  pages={1--5},
  year={1987},
  publisher={Elsevier}
}

@article{lotti2020adaptive,
  title={Adaptive model-based myoelectric control for a soft wearable arm exosuit: A new generation of wearable robot control},
  author={Lotti, Nicola and Xiloyannis, Michele and Durandau, Guillaume and Galofaro, Elisa and Sanguineti, Vittorio and Masia, Lorenzo and Sartori, Massimo},
  journal={IEEE Robotics \& Automation Magazine},
  volume={27},
  number={1},
  pages={43--53},
  year={2020},
  publisher={IEEE}
}

@article{schonhaut2025emg,
  title={Is EMG Information Necessary for Deep Learning Estimation of Joint and Muscle Level States?},
  author={Schonhaut, Ethan B and Scherpereel, Keaton L and Young, Aaron J},
  journal={IEEE Transactions on Biomedical Engineering},
  year={2025},
  publisher={IEEE}
}

@article{lenzi2012intention,
  title={Intention-based EMG control for powered exoskeletons},
  author={Lenzi, Tommaso and De Rossi, Stefano Marco Maria and Vitiello, Nicola and Carrozza, Maria Chiara},
  journal={IEEE transactions on biomedical engineering},
  volume={59},
  number={8},
  pages={2180--2190},
  year={2012},
  publisher={IEEE}
}

@inproceedings{polygerinos2015emg,
  title={EMG controlled soft robotic glove for assistance during activities of daily living},
  author={Polygerinos, Panagiotis and Galloway, Kevin C and Sanan, Siddharth and Herman, Maxwell and Walsh, Conor J},
  booktitle={2015 IEEE international conference on rehabilitation robotics (ICORR)},
  pages={55--60},
  year={2015},
  organization={IEEE}
}

@article{ferroni2025soft,
  title={A soft pneumatic exosuit to assist pronosupination in individuals with spinal cord injury},
  author={Ferroni, Roberto and D’Avola, Gaetano and Mauceri, Daniele Filippo and Pau, Chiara and Sciarrone, Giorgia and Righi, Gabriele and Carpaneto, Jacopo and Gandolla, Marta and Del Popolo, Giulio and Micera, Silvestro and others},
  journal={Advanced Intelligent Systems},
  volume={7},
  number={12},
  pages={e202500124},
  year={2025},
  publisher={Wiley Online Library}
}

@article{realmuto2022assisting,
  title={Assisting Forearm Function in Children With Movement Disorders via A Soft Wearable Robot With Equilibrium-Point Control},
  author={Realmuto, Jonathan and Sanger, Terence D},
  journal={Frontiers in Robotics and AI},
  volume={9},
  pages={877041},
  year={2022},
  publisher={Frontiers Media SA}
}

@article{zhou2021kinematics,
  title={Kinematics-based control of an inflatable soft wearable robot for assisting the shoulder of industrial workers},
  author={Zhou, Yu Meng and Hohimer, Cameron and Proietti, Tommaso and O’Neill, Ciar{\'a}n Tom{\'a}s and Walsh, Conor J},
  journal={IEEE Robotics and Automation Letters},
  volume={6},
  number={2},
  pages={2155--2162},
  year={2021},
  publisher={IEEE}
}

@article{proietti2023restoring,
  title={Restoring arm function with a soft robotic wearable for individuals with amyotrophic lateral sclerosis},
  author={Proietti, Tommaso and O’Neill, Ciaran and Gerez, Lucas and Cole, Tazzy and Mendelowitz, Sarah and Nuckols, Kristin and Hohimer, Cameron and Lin, David and Paganoni, Sabrina and Walsh, Conor},
  journal={Science Translational Medicine},
  volume={15},
  number={681},
  pages={eadd1504},
  year={2023},
  publisher={American Association for the Advancement of Science}
}

@article{xiloyannis2019physiological,
  title={Physiological and kinematic effects of a soft exosuit on arm movements},
  author={Xiloyannis, Michele and Chiaradia, Domenico and Frisoli, Antonio and Masia, Lorenzo},
  journal={Journal of neuroengineering and rehabilitation},
  volume={16},
  number={1},
  pages={29},
  year={2019},
  publisher={Springer}
}

@article{chiaradia2021assistive,
  title={An assistive soft wrist exosuit for flexion movements with an ergonomic reinforced glove},
  author={Chiaradia, Domenico and Tiseni, Luca and Xiloyannis, Michele and Solazzi, Massimiliano and Masia, Lorenzo and Frisoli, Antonio},
  journal={Frontiers in Robotics and AI},
  volume={7},
  pages={595862},
  year={2021},
  publisher={Frontiers Media SA}
}

@article{young2017biomechanical,
  title={A biomechanical comparison of proportional electromyography control to biological torque control using a powered hip exoskeleton},
  author={Young, Aaron J and Gannon, Hannah and Ferris, Daniel P},
  journal={Frontiers in bioengineering and biotechnology},
  volume={5},
  pages={37},
  year={2017},
  publisher={Frontiers Media SA}
}

\end{document}